\documentclass[10pt,twocolumn,letterpaper]{article}

\usepackage{cvpr}
\usepackage{times}
\usepackage{epsfig}
\usepackage{graphicx}
\usepackage{amsmath}
\usepackage{amssymb}

\usepackage{tabularx}
\usepackage{subfigure}
\usepackage{booktabs}
\usepackage{multirow}
\usepackage[super]{nth}
\usepackage[linesnumbered,ruled,lined,commentsnumbered]{algorithm2e}

\usepackage[pagebackref=true,breaklinks=true,letterpaper=true,colorlinks,bookmarks=false]{hyperref}

\cvprfinalcopy 

\def\cvprPaperID{7235} 

\ifcvprfinal\fi
\begin{document}

\title{Scene-Adaptive Video Frame Interpolation via Meta-Learning}

\author{Myungsub Choi\textsuperscript{1}\hspace*{-0.075cm}\\
\textsuperscript{1}ASRI, Department of ECE, Seoul National University ~~~~ \textsuperscript{2}Department of CS, Hanyang University\hspace*{-14cm}\\
\textsuperscript{1}{\tt\small \{cms6539, ultio791, dsybaik, kyoungmu\}@snu.ac.kr} ~~~~ \textsuperscript{2}{\tt\small taehyunkim@hanyang.ac.kr}\hspace*{-14cm}
\and
Janghoon Choi\textsuperscript{1}\hspace*{-0.075cm}
\and
Sungyong Baik\textsuperscript{1}\hspace*{-0.075cm}
\and
Tae Hyun Kim\textsuperscript{2}\hspace*{-0.075cm}
\and
Kyoung Mu Lee\textsuperscript{1}
}

\maketitle

\begin{abstract}

Video frame interpolation is a challenging problem because there are different scenarios for each video depending on the variety of foreground and background motion, frame rate, and occlusion.
It is therefore difficult for a single network with fixed parameters to generalize across different videos.
Ideally, one could have a different network for each scenario, but this is computationally infeasible for practical applications.
In this work, we propose to adapt the model to each video by making use of additional information that is readily available at test time and yet has not been exploited in previous works.
We first show the benefits of `test-time adaptation' through simple fine-tuning of a network, then we greatly improve its efficiency by incorporating meta-learning.
We obtain significant performance gains with only a single gradient update without any additional parameters.
Finally, we show that our meta-learning framework can be easily employed to any video frame interpolation network and can consistently improve its performance on multiple benchmark datasets.

\end{abstract}

\section{Introduction}
\label{sec:introduction}


Video frame interpolation aims to upscale the temporal resolution of a video, by synthesizing intermediate frames in-between the neighboring frames of the original input.
Owing to its wide range of applications, including slow-motion generation and frame-rate up-conversion that provide better visual experiences with more details and less motion blur, video frame interpolation has gained substantial interest in the computer vision community.
Recent advances of deep convolutional neural networks (CNNs) for video frame interpolation~\cite{jiang2018superslomo,liu2017dvf,niklaus2018cas,niklaus2017adaconv,niklaus2017sepconv,xue2018toflow} lead to a significant boost in performance.
However, generating high-quality frames is still a challenging problem due to large motion and occlusion in a diverse set of scenes.


Previous approaches to video frame interpolation~\cite{jiang2018superslomo,liu2017dvf,niklaus2018cas,niklaus2017adaconv,niklaus2017sepconv,xue2018toflow}, as well as other learning-based video processing models~\cite{carreira2017quovadis,cheng2017segflow,simonyan2014two-stream,zhu2018CVPR_video_det,zhu2017flow}, typically require a huge amount of data for training.
However, videos in the wild comprise of various distinctive scenes with many different types of low-level patterns.
This makes it difficult for a single model to perform well on all possible test cases, even if trained with large datasets.

\begin{figure}[t]
	\begin{center}
		\includegraphics[width=0.95\linewidth]{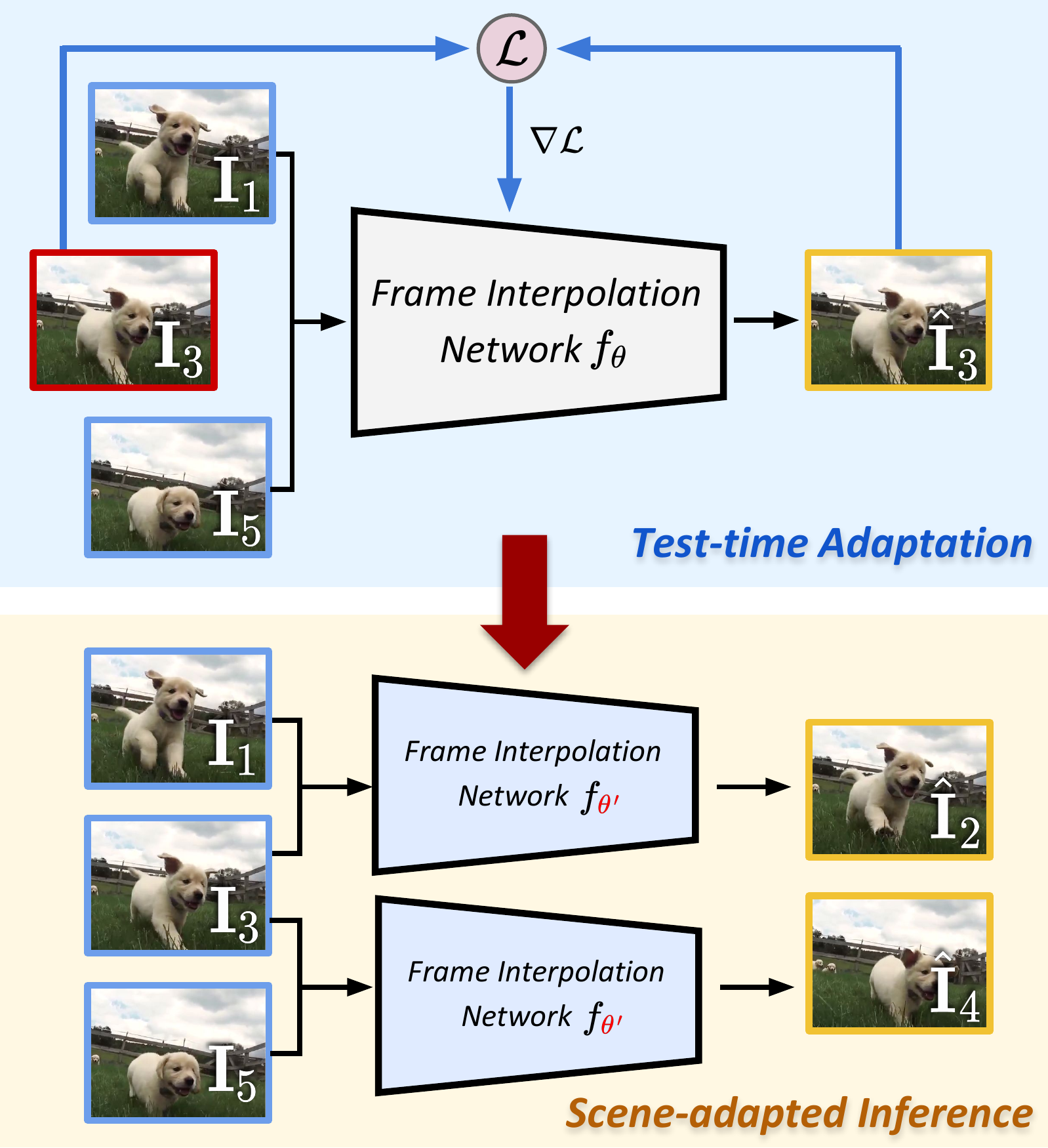}
	\end{center}
	\vspace{-2mm}
	\caption{{\textbf{Motivation of the proposed video frame interpolation method.}} Our video frame interpolation framework incorporates a test-time adaptation process followed by scene-adapted inference. The adaptation process takes advantage of additional information from the input frames and is quickly performed with only a single gradient update to the network.}
	\label{fig:intro}
	\vspace{-3mm}
\end{figure}


This problem can be alleviated by making the model adaptive to the specific input data.
Utilizing the additional information only available at test time and customizing the model to each of the test data samples has shown to be effective in numerous areas.
Examples include single-image super-resolution approaches exploiting self-similarities inherent in the target image~\cite{glasner2009super,huang2015single,huang2017srhrf+,michaeli2013nonparametric,shocher2018zero}, or many visual tracking methods where online adaptation to the input video sequence is crucial in performance~\cite{choi2017deep,danelljan2017eco,nam2016learning}.
However, most works either increase the number of parameters or require considerable inference time for test-time adaptation of the network parameters.


Meta-learning, also known as \textit{learning to learn}, can take a step forward to remedy current limitations in test-time adaptation. 
The goal of meta-learning is to design algorithms or models that can quickly adapt to new tasks from small set of training examples given during testing phase.
It has been gaining tremendous interest in solving few-shot classification/regression problems as well as some reinforcement learning applications~\cite{finn2017model}, but employing meta-learning techniques to low-level computer vision problems has yet to be explored.

To this end, we propose a scene-adaptive video frame interpolation algorithm that can rapidly adapt to new, unseen videos (or tasks, in meta-learning viewpoint) at test time and achieve substantial performance gain.
A brief overview of the main idea of our approach is illustrated in Fig.~\ref{fig:intro}.
Using any off-the-shelf existing video frame interpolation framework, our algorithm updates its parameters using the frames only available at test time, and uses the adapted model to interpolate intermediate frames in the same way as the conventional approaches.
Although the proposed method is not applicable for videos with their total length of less than 3 frames, most real-world scenarios have multiple consecutive frames that we can fully utilize for our meta-learning based test-time adaptation scheme.

Overall, our contributions are summarized as follows:
\begin{itemize}
\setlength\itemsep{1pt}
\item We propose a novel adaptation framework that can further improve conventional frame interpolation models without changing their architectures.
\item To the best of our knowledge, the proposed approach is the first integration of meta-learning techniques for test-time adaptation in video frame interpolation. 
\item We confirm that our framework consistently improves upon even the most recent state-of-the-art methods.
\end{itemize}


\section{Related works}
\label{sec:related}

In this section, we review the extensive literature of video frame interpolation.
Existing test-time adaptation schemes for other low-level vision applications and the history of meta-learning algorithms are also described.

\vspace{-3mm}
\paragraph{Video frame interpolation: }
While video frame interpolation has a long-established history, we concentrate on recent learning-based algorithms, particularly CNN-based interpolation approaches.

The first attempt to incorporate CNNs to video frame interpolation was done by Long~\etal~\cite{long2016learning}, where interpolation is obtained as a byproduct of self-supervised learning of optical flow estimation.
Since then, numerous approaches have focused on effectively modeling motion and handling occlusions.
Meyer~\etal~\cite{meyer2018phasenet,meyer2015phase} represent motion as per-pixel phase shift, and Niklaus~\etal~\cite{niklaus2017adaconv,niklaus2017sepconv} model the sequential process of motion estimation and frame synthesis into a single spatially-adaptive convolution step.
Choi~\etal~\cite{choi2020cain} handles motion with a simple feedforward network with channel attention.

Another line of research use optical flow estimation as an intermediate step (as a proxy) and warp the original frames with the estimated motion map for alignment, followed by further refinement and occlusion handling to obtain the final interpolations~\cite{bao2019DAIN,bao2018memc,jiang2018superslomo,liu2019cyclicgen,liu2017dvf,niklaus2018cas,xu2019quadratic,xue2018toflow}.
These flow-based methods are generally able to synthesize sharp and natural frames, but some heavily depend on the pre-trained optical flow estimation network and show doubling artifacts in cases with large motion when flow estimation fail.
Recently, Bao~\etal~\cite{bao2019DAIN} additionally use depth map estimation model to compensate for the missing information in flow estimation and effectively handle the occluding regions.

\vspace{-3mm}
\paragraph{Test-time adaptation: }
Contrary to previous works, we explore an orthogonal area of research, adaptation to the inputs at test time, to further improve the accuracy of given video frame interpolation models.
Our work is inspired by the success of self-similarity based approaches in image super-resolution~\cite{glasner2009super,huang2015single,huang2017srhrf+,michaeli2013nonparametric,shocher2018zero}.
Notably, recent zero-shot super-resolution (ZSSR) method proposed by Shocher~\etal~\cite{shocher2018zero} has shown impressive results by incorporating deep learning.
Specifically, ZSSR at test time extracts the patches only from the input image and trains a small image-specific CNN, thereby naturally exploiting the information that is only available after observing the test inputs.
However, ZSSR suffers from slow inference time due to its self-training step, and it is prone to overfitting since using a pretrained network trained with large external datasets is not viable for internal training.

For video frame interpolation, Reda~\etal~\cite{reda2019unsupervised} recently proposed the first approach to adapt to the test data in an unsupervised manner by using a cycle-consistency constraint.
However, their method adapts to the general domain of the test data, and cannot adapt to each test sample.
On the other hand, the proposed algorithm enables to update the model parameters \textit{w.r.t.} each local part of the test sequence, thus better adapting to local motions and scene textures.

\vspace{-3mm}
\paragraph{Meta-learning: }
To achieve test-time adaptation without susceptibility to overfitting and without greatly increasing the cost of computation, we turn our attention to meta-learning. 
Recently, meta-learning has gained a lot of attention for its high performance in few-shot classification, which evaluates the capability of the system to adapt to new classification tasks with few examples. 
Meta-learning aims to achieve such adaptation to new tasks (videos in our case) through learning prior knowledge across tasks. \cite{bengio1992optimization, hochreiter2001learning, schmidhuber1987evolutionary, schmidhuber1992learning, thrun2012learning}. 
Broadly, one can categorize meta-learning systems into three classes: metric-based, network-based, and optimization-based. 
The metric-based meta-learning manifests the prior knowledge by learning a feature embedding space, where different classes are placed far apart and similar classes are placed close to each other \cite{koch2015siamese, NIPS2017_6996, Sung_2018_CVPR, vinyals2016matching}. 
The learned embedding space is then used to learn relationship between a query and support examples in few-shot classification. 
Network-based meta-learning achieves fast adaptation through encoding input-dependent dynamics into the architecture itself by generating input-conditioned weights \cite{munkhdalai2017meta, oreshkin2018tadam} or employing an external memory \cite{munkhdalai2018rapid, santoro2016meta}. 
On the other hand, optimization-based systems aim to encode the prior knowledge into optimization process for fast adaptation~\cite{finn2017model,nichol2018first,ravi2017optimization}.
Among optimization-based systems, MAML~\cite{finn2017model} has greatly enjoyed the attention for its simplicity and generalizability, in contrast to the metric or network-based systems that suffer from the limitations in either applications or scalability issues.
The generalizability of its model-agnostic algorithm motivates us to use MAML to integrate test-time adaptation into video frame interpolation. 

\section{Proposed Method}
\label{sec:model}

In this section, we first describe the general problem settings for video frame interpolation.
Then, we empirically show the advantage of test-time adaptation with a feasibility test, and justify the need for meta-learning in this scenario.

\subsection{Video frame interpolation problem set-up}

The goal of video frame interpolation algorithms is to generate a high-quality, high frame-rate video given a low frame-rate input video by synthesizing intermediate frames between two neighboring frames.
Standard settings for most frame interpolation models receive two input frames and output a single intermediate frame.
Specifically, if we let $\mathbf{I}_1$ and $\mathbf{I}_3$ be the two consecutive input frames, our goal is to synthesize the middle frame $\hat{\mathbf{I}}_2$.
Although recent frame interpolation models also consider more complex multi-frame interpolation problem where a frame of any arbitrary time step between two frames can be synthesized, we constrain our discussions to the single-frame interpolation models in this work.
However, note that our proposed meta-learning framework described in Sec.~\ref{sec:meta} is model-agnostic and easily generalizable to different settings as long as the model is differentiable.


\subsection{Exploiting extra information at test time}
\label{sec:feasibility}

\begin{figure}[t]
	\begin{center}
		\includegraphics[width=1.0\linewidth]{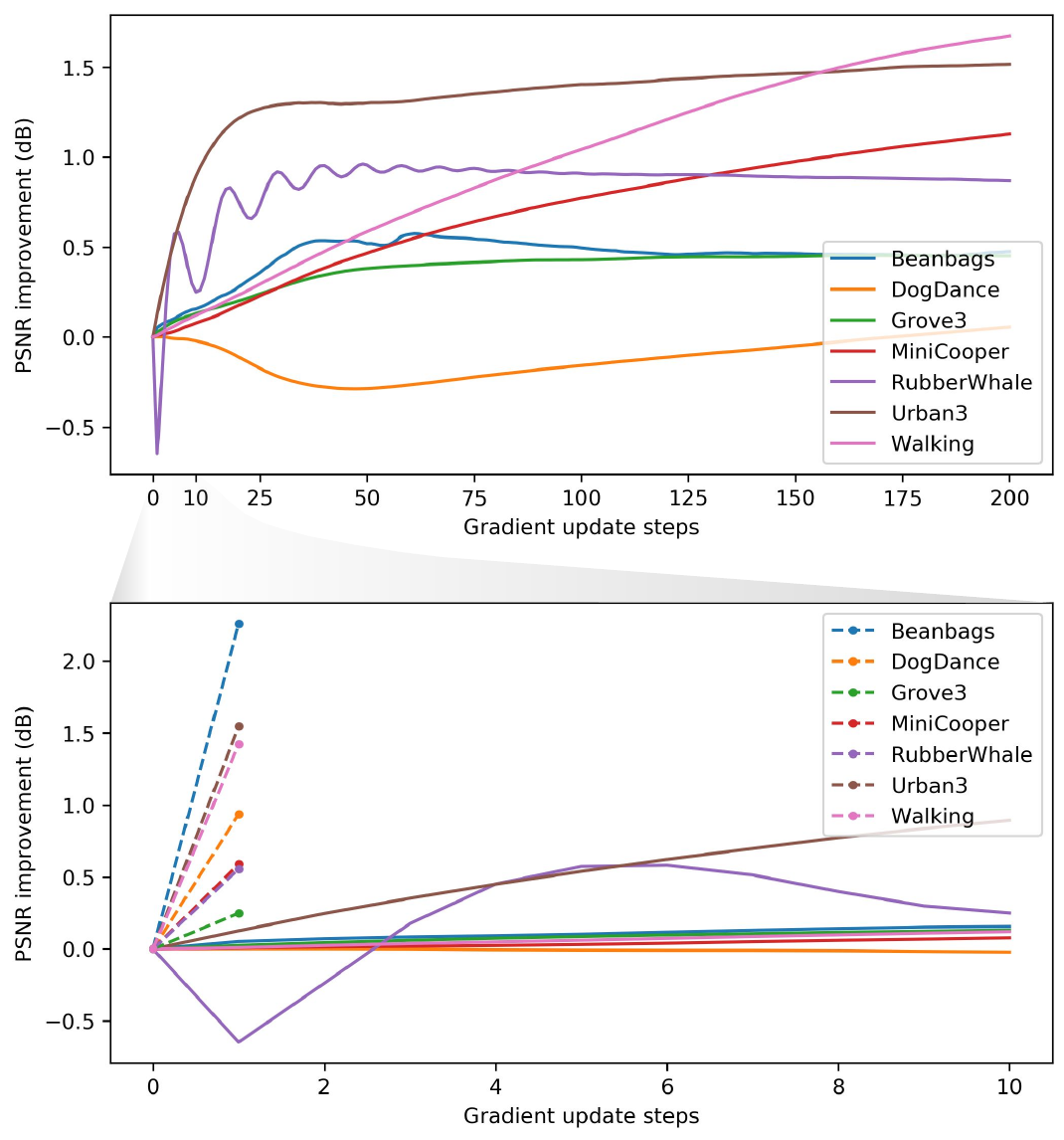}
	\end{center}
	\vspace{-3mm}
	\caption{{\textbf{Feasibility test for test-time adaptation.}} Upper graph shows that fine-tuning with the test input data can improve performance in general, but the number of required steps greatly differs for each sequence.
	Lower graph shows a $\times20$ zoomed in version of the upper graph, additionally denoting the large performance gain obtained with our \textit{meta-learned} SepConv with a single gradient update.}
	\label{fig:feasibility}
	\vspace{-5mm}
\end{figure}

\begin{figure*}[t]
	\begin{center}
		\includegraphics[width=1.0\linewidth]{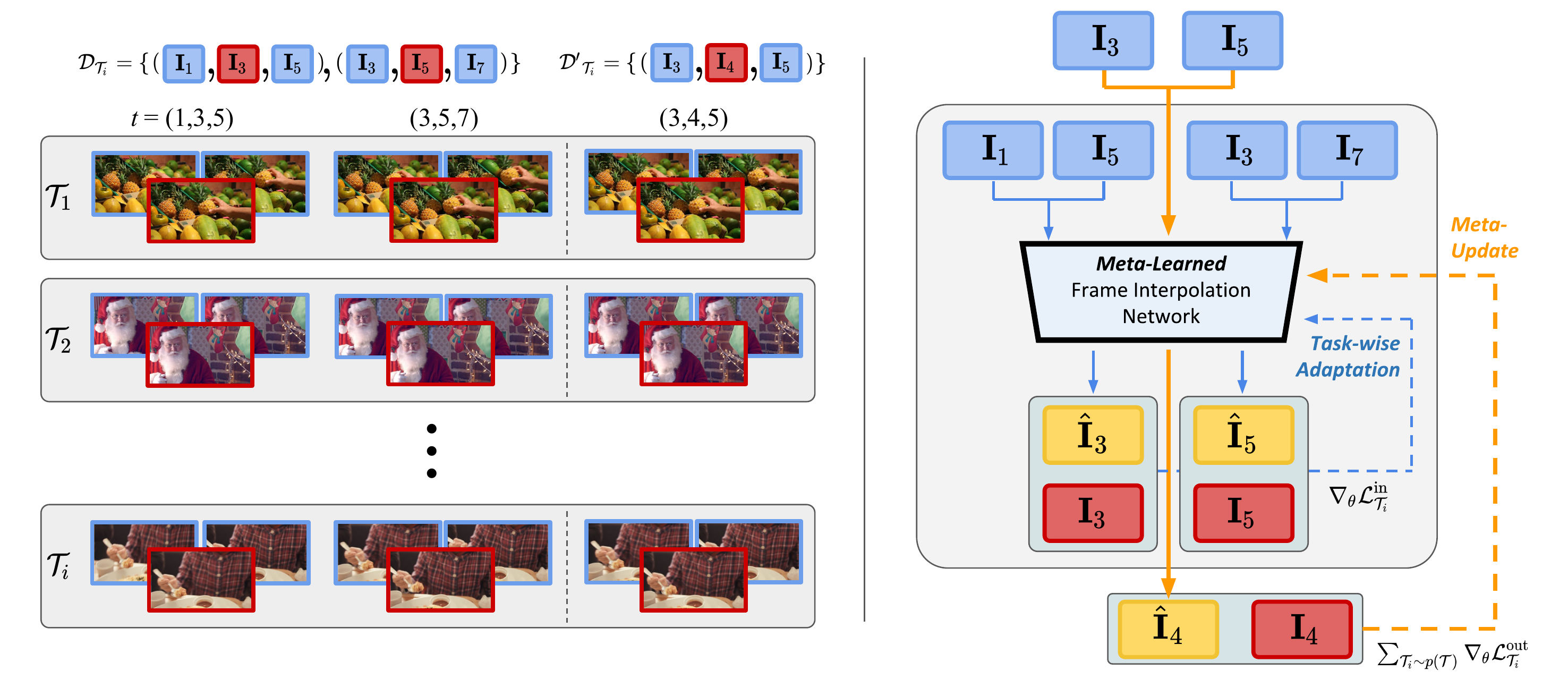}
	\end{center}
	\vspace{-0mm}
	\caption{{\textbf{Overview of the training process for the proposed video frame interpolation network.}} 
	\textit{\textbf{Left}}: Each task $\mathcal{T}_i$ consists of three frame triplets chosen from a video sequence where two are used for task-wise adaptation (\textit{i.e.}, inner loop update) and one is used for meta-update (\textit{i.e.}, outer loop update).
	\textit{\textbf{Right}}: Network parameters $\theta$ are adapted by gradient descent on loss $\mathcal{L}^{\text{in}}_{\mathcal{T}_i}$ using triplets in $\mathcal{D}_{\mathcal{T}_i}$ and stored for each task, and meta-update is performed by minimizing the sum of each loss $\mathcal{L}^{\text{out}}_{\mathcal{T}_i}$ using the triplets in $\mathcal{D'}_{\mathcal{T}_i}$ for all tasks.}
	\label{fig:overview}
	\vspace{-0mm}
\end{figure*}

We demonstrate the effectiveness of test-time adaptation with a feasibility test and describe the details on our design choices.
Starting from a baseline pre-trained frame interpolation model, we aim to fine-tune the model parameters at test time to improve its performance (for each test video sequence).
To fine-tune the model, a frame triplet consisting of 3 consecutive frames are needed, where the first and last frames become the input and the middle frame becomes the target output.
While training (fine-tuning) with triplets of a low frame-rate video may seem not beneficial due to the wider time gap, the overall interpolation performance boost has been observed, as shown in the following experiment. This implies the importance of the context and attributes of the given video, such as unique motion and occlusion, and signifies the benefit of test-time adaptation.

For a feasibility test on the effectiveness of test-time adaptation, we fine-tune a pre-trained SepConv~\cite{niklaus2017sepconv} model on each sequence from Middlebury~\cite{baker2010flow} dataset.
Specifically, we choose 7 sequences from \texttt{OTHERS} set, and fine-tune the baseline model with Adamax~\cite{kingma2014Adam} optimizer (which was used to train the original SepConv model) with a fixed learning rate of $10^{-5}$.
Batch construction for the gradient update is analogous to Fig.~\ref{fig:intro}, but we increase the number of frames for test-time adaptation from 3 ($t=1,3,5$) to 4 ($t=1,3,5,7$).
In a sense, it can be seen as a 2-shot update, since we can build 2 triplets ($t=(1, 3, 5), (3, 5, 7)$) from the 4 input frames.
Updating the model parameters with these 2 triplets for many iterations can tell whether or not this test-time adaptation scheme is advantageous.
We measure the performance with peak signal-to-noise ratio (PSNR), and the results for PSNR difference with respect to the number of gradient update steps are shown in Fig.~\ref{fig:feasibility}.

The characteristics for performance improvements, shown in the upper graph of Fig.~\ref{fig:feasibility}, greatly differs from sequence to sequence.
While the PSNR scores for \textit{Minicooper} and \textit{Walking} steadily improve for 200 gradient updates and do not overfit even after over 1dB gain, updating with \textit{DogDance} sequence hurts the original model's performance in its early stage. 
Notably, the graph for \textit{RubberWhale} shows a strange characteristic, where the performance severely drops after the first gradient update but suddenly shifts back to the positive side after the subsequent steps.
From these results, we can arguably conclude that test-time adaptation is beneficial for video frame interpolation, but how much to adapt (or not adapt at all to avoid overfitting) for each different sequence is hard to decide.

By incorporating meta-learning techniques, our method can enhance the original SepConv model to rapidly adapt to the test sequence, without changing any architectural choices or introducing additional parameters.
With just a single gradient update at test time, our \textit{meta-learned} SepConv can achieve large performance gain, as illustrated in the lower graph of Fig.~\ref{fig:feasibility}.
Compared to hundreds of iterations required for fine-tuning the baseline model, our meta-learned SepConv extremely reduces the computation time needed to obtain the same amount of performance boost.


\subsection{Background on MAML}
Meta-learning aims at rapidly adapting to novel tasks with only a few examples \textit{i.e.} few-shot learning.
Recent model-agnostic meta-learning (MAML)~\cite{finn2017model} approach achieve this goal with only a few gradient update iterations by preparing the model to be readily adaptable to incoming test data.
In other words, MAML finds a good initialization of the parameters that are sensitive to changes in task, so that small updates can make large improvements on reducing the error measures and boosting the performance for each new task.
Before diving into the main algorithm, we would first like to start with the formulation of the general meta-learning and MAML.

Under the assumption of the existence of task distribution, $p(\mathcal{T})$, the goal of MAML is to learn the initialization parameters that represent the prior knowledge that exists throughout the task distribution. 
In $k$-shot learning setting, $\mathcal{D}_{\mathcal{T}_i}$, a set of $k$ number of examples, are sampled from each task, $\mathcal{T}_i \sim p(\mathcal{T})$. 
The sampled examples, along with its corresponding loss $\mathcal{L}_{\mathcal{T}_i}$, roughly represent the task itself and are used for the model to adapt to the task. In MAML, this is achieved by fine-tuning:
\begin{equation}
\theta'_i = \theta - \alpha\nabla_{\theta} \mathcal{L}_{\mathcal{T}_i}(f_\theta).
\label{eq:inner_update}
\end{equation}
Once the model is adapted to each task, $\mathcal{T}_i$, new examples, $\mathcal{D}'_{\mathcal{T}_i}$, are sampled from the same task to evaluate the generalization of the adapted model on unseen examples. 
The evaluation acts as a feedback for MAML to adjust its initialization parameters to achieve better generalization:
\begin{equation}
\theta  \leftarrow \theta - \beta \nabla_{\theta} \sum_{\mathcal{T}_i}{\mathcal{L}_{\mathcal{T}_i}(f_{\theta'_i})}.
\end{equation}

\subsection{Meta-learning for frame interpolation}
\label{sec:meta}


For video frame interpolation, we define a \textit{task} as performing frame interpolation on a frame sequence (video).
Fast adaptation to new video scenes via MAML introduces our scene-adaptive frame interpolation algorithm, which is described in detail later in this section.


We consider a frame interpolation model $f_{\theta}$, parameterized by $\theta$, that receives two input frames $(\mathbf{I}_t, \mathbf{I}_{t+2T})$ and outputs the estimated middle frame $\hat{\mathbf{I}}_{t+T}$ for any time step $t$ and interval $T$.
Thus, a training sample needed to update the model parameters can be formalized as a frame triplet  $( \mathbf{I}_t, \mathbf{I}_{t+T}, \mathbf{I}_{t+2T})$.
We define a task $\mathcal{T}$ as minimizing the sum of the losses $\mathcal{L}:\{(\mathbf{I}_t, \mathbf{I}_{t+T}, \mathbf{I}_{t+2T})\} \rightarrow \mathbb{R}$ for all time steps $t$ in low frame-rate input video.
In our scene-adaptive frame interpolation setting, each new task $\mathcal{T}_i$ drawn from $p(\mathcal{T})$ consists of frames in a single sequence, and the model is adapted to the task using a task-wise training set $\mathcal{D}_{\mathcal{T}_i}$, where training triplets are constructed only with frames existent in the low frame-rate input.
Updating parameters at meta-training stage is governed by the loss $\mathcal{L}^{\text{out}}_{\mathcal{T}_i}$ for a task-wise test set $\mathcal{D}'_{\mathcal{T}_i}$, where the test triplets consist of two input frames and the target ground-truth intermediate frame that is non-existent in the low frame-rate input.
In practice, we use 4 input frames $\{ \mathbf{I}_1, \mathbf{I}_3, \mathbf{I}_5, \mathbf{I}_7 \}$ as described in Sec.~\ref{sec:feasibility}, and 1 target middle frame $\mathbf{I}_4$.
The task-wise training and test set then become $\mathcal{D}_{\mathcal{T}_i} = \{ (\mathbf{I}_1, \mathbf{I}_3, \mathbf{I}_5),( \mathbf{I}_3, \mathbf{I}_5, \mathbf{I}_7) \}$ and $\mathcal{D}'_{\mathcal{T}_i} = \{ (\mathbf{I}_3, \mathbf{I}_4, \mathbf{I}_5) \}$.
These configurations are illustrated in the left part of Fig.~\ref{fig:overview}.


Given the above notations, we now describe the flow of our scene-adaptive frame interpolation algorithm in more detail.
Since our method is model-agnostic due to integration with MAML, we can use any existing video frame interpolation model as a baseline.
However, unlike MAML where the model parameters begin from random initialization, we initialize the model parameters from a pre-trained model that is already capable of generating sensible interpolations.
Thus, our algorithm can also be viewed as a post-processing step, where the baseline model is updated to be readily adaptive to each test video for further performance boost.

The detailed flow of the algorithm is illustrated in the right part of Fig.~\ref{fig:overview}.
Let us denote the update iterations for each task as \textit{inner loop} and the meta-update iterations as \textit{outer loop}.
For inner loop training, given two frame triplets from task-wise training set $\mathcal{D}_{\mathcal{T}_i}$ for each task $\mathcal{T}_i$, we first calculate the model predictions as 
\begin{equation}
\label{eq:I3I5}
    \hat{\mathbf{I}}_3 = f_{\theta}(\mathbf{I}_1, \mathbf{I}_5), ~~~~~\hat{\mathbf{I}}_5 = f_{\theta}(\mathbf{I}_3, \mathbf{I}_7),
\end{equation}
where the superscript $i$ is hidden to reduce notation clutter.
These outputs are then used to compute the loss for inner loop update $\mathcal{L}_{\mathcal{T}_i}^{\text{in}}(f_{\theta})$, calculated as the sum of two losses as in 
\begin{equation}
\label{eq:loss_in}
    \mathcal{L}_{\mathcal{T}_i}^{\text{in}}(f_{\theta}) = \mathcal{L}_{\mathcal{T}_i}(\hat{\mathbf{I}}_3, \mathbf{I}_3) + \mathcal{L}_{\mathcal{T}_i}(\hat{\mathbf{I}}_5, \mathbf{I}_5).
\end{equation}
Next, we calculate the gradients for $\mathcal{L}_{\mathcal{T}_i}^{\text{in}}(f_{\theta})$ and update $\theta$ with gradient descent to obtain customized parameters $\theta'_i$ for each task $\mathcal{T}_i$.
Note that we can use any gradient-based optimizer (\eg Adam~\cite{kingma2014Adam}) for the updating step, and we choose the same optimization algorithm used to train the baseline pre-trained model in practice.
Also note that the inner loop update can optionally consist of multiple iterations such that $\theta'_i$ is a result of $k$ gradient updates from $\theta$, where $k$ is the number of iterations. 
We analyze the effect of hyperparameter $k$ in Sec.~\ref{sec:ablation}, and choose $k=1$ throughout our experiments for performance and simplicity (see Table~\ref{tb:ablation_inner_updates}).
To further reduce computation, we employ a first-order approximation as suggested in~\cite{finn2017model} and avoid calculating the second-order derivatives required for the nested-loop updates in meta-training.

When training the outer loop, the parameters are updated to minimize the losses for $f_{\theta'_i}$ with respect to $\theta$, on each of the task-wise test triplet $\{  ( \mathbf{I}_3, \mathbf{I}_4, \mathbf{I}_5)  \} \in \mathcal{D}'_{\mathcal{T}_i}$.
Loss function for the outer loop meta-update is defined as
\begin{equation}
\label{eq:loss_out}
    \mathcal{L}_{\mathcal{T}_i}^{\text{out}}(f_{\theta'_i}) = \mathcal{L}_{\mathcal{T}_i}(f_{\theta'_i}(\mathbf{I}_3, \mathbf{I}_5), \mathbf{I}_4),
\end{equation}
and the summation of all losses for the sampled batch of sequences (tasks) $\mathcal{T}_i \sim p(\mathcal{T})$ are used to calculate the gradient and update the model parameters.
The overall training process is summarized in Algorithm~\ref{alg:meta_train}.

\begin{algorithm}[t]
	\SetKwData{Left}{left}\SetKwData{This}{this}\SetKwData{Up}{up}
	\SetKwFunction{Union}{Union}\SetKwFunction{FindCompress}{FindCompress}
	\SetKwInOut{Require}{Require}
	\SetAlgoLined
	
	\Require{$p(\mathcal{T})$: uniform distribution over sequences}
	\Require{$\alpha, \beta$: step size hyper-parameters}
	\BlankLine
	
	Initialize parameters $\theta$ \\
	\While{not converged}{
	    Sample batch of sequences $\mathcal{T}_i \sim p(\mathcal{T})$ \\
	    \ForEach{i}{
	        Generate triplets $\mathcal{D}_{\mathcal{T}_i} = \{  ( \mathbf{I}_1, \mathbf{I}_3, \mathbf{I}_5),( \mathbf{I}_3, \mathbf{I}_5, \mathbf{I}_7)  \}$ from $\mathcal{T}_i$ \\
	        Compute $\hat{\mathbf{I}}_3$, $\hat{\mathbf{I}}_5$ in Eq.~\eqref{eq:I3I5}\\
	        Evaluate $\nabla_{\theta} \mathcal{L}^{\text{in}}_{\mathcal{T}_i}(f_\theta)$ using $\mathcal{L}_{\mathcal{T}_i}$ in Eq.~\eqref{eq:loss_in}\\
		    Compute adapted parameters with gradient descent: $\theta_i' = \theta - \alpha\nabla_{\theta}\mathcal{L}^{\text{in}}_{\mathcal{T}_i}(f_\theta)$\\
		    Generate and save triplet $\mathcal{D}'_{\mathcal{T}_i} = \{  ( \mathbf{I}_3, \mathbf{I}_4, \mathbf{I}_5)  \}$ from $\mathcal{T}_i$ for the meta-update \\
	    }
	    Update $\theta \leftarrow \theta - \beta\nabla_{\theta}\sum_{\mathcal{T}_i \sim p(\mathcal{T})}\mathcal{L}^{\text{out}}_{\mathcal{T}_i}(f_{\theta_i'})$ using each $\mathcal{D}_{\mathcal{T}_i}'$ and $\mathcal{L}_{\mathcal{T}_i}$ in Eq.~\eqref{eq:loss_out}\\
	}
	\caption{Scene-Adaptive Frame Interpolation} \label{alg:meta_train}
\end{algorithm}

At test time, the base parameters $\theta$ for the outer loop are fixed, and only the inner loop update is performed to modify the parameter values to $\theta'_i$ for each test sequence $\mathcal{T}_i$.
The final interpolations can then be obtained as the output of the adapted model $f_{\theta'_i}$.

\begin{table*}
	\centering
	\caption{\textbf{Quantitative results for meta-training for recent frame interpolation algorithms}.
	    We evaluate the benefits of our scene-adaptive algorithm on 3 datasets: VimeoSeptuplet~\cite{xue2018toflow}, Middlebury-\texttt{OTHERS}~\cite{baker2010flow}, and HD~\cite{bao2018memc} dataset.
	    Performance is measured in PSNR (dB). Note how our \textit{Meta-trained} performance consistently improves upon the \textit{Baseline} or \textit{Re-trained} correspondents.}
	\vspace{0.2cm}
	\scalebox{0.85}{
		\begin{tabular}{l |c c c | c c c | c c c}
			& \multicolumn{3}{c}{VimeoSeptuplet~\cite{xue2018toflow}} & \multicolumn{3}{c}{Middlebury-\texttt{OTHERS}~\cite{baker2010flow}} & \multicolumn{3}{c}{HD~\cite{bao2018memc}} \\
			\toprule
			Method & Baseline & Re-trained & Meta-trained & Baseline & Re-trained & Meta-trained & Baseline & Re-trained & Meta-trained\\
			\midrule
			DVF~\cite{liu2017dvf} & 26.60 & 32.21 & \textbf{32.27} & 26.70 & 29.51 & \textbf{29.70} & --- & --- & --- \\ 
			SuperSloMo~\cite{jiang2018superslomo} & 30.85 & 32.76 & \textbf{33.12} & 30.28 & 33.54 & \textbf{33.70} & 26.05 & 29.66 & \textbf{29.81} \\
			SepConv~\cite{niklaus2017sepconv} & 33.70 & 33.72 & \textbf{34.17} & 35.14 & 34.90 & \textbf{35.81} & 30.04 & 30.01 & \textbf{30.19} \\
			DAIN~\cite{bao2019DAIN} & 34.73 & 34.86 & \textbf{34.94} & \textbf{36.57} & 36.50 & 36.50 & 30.35 & 30.45 & \textbf{30.51} \\
			\bottomrule
		\end{tabular}
	}
	\label{tb:quantitative}
	\vspace{-3mm}
\end{table*}

Note that, the biggest difference from our algorithm from the original MAML is that the distributions for the task-wise training and test set, $\mathcal{D}_{\mathcal{T}_i}$ and $\mathcal{D}'_{\mathcal{T}_i}$, are not the same.
Namely, $\mathcal{D}_{\mathcal{T}_i}$ have a broader spectrum of motion and includes $\mathcal{D}'_{\mathcal{T}_i}$, since the time gap between the frame triplets are twice as large.
Though this case with a distribution gap is an unexplored area in meta-learning literature, it shows an encouraging effect for the task of video frame interpolation; the model trained with our algorithm learns to update itself in considerably more difficult scenarios with larger motion, learning the overall context and motion present in the video as a result.
Interpolations for the original input frames then become an easy task for our well-adapted model, which results in performance gain.
Both quantitative and qualitative results in the experiments show that our algorithm actually improves the original model to better handle bigger motion.
\section{Experiments}
\label{sec:experiments}

\subsection{Settings}
\label{sec:experiment_settings}

\paragraph{Datasets}

Most of the existing works on video frame interpolation use the video data pre-processed into frame triplets.
Though our baseline model is pre-trained with conventional triplet datasets, it is not applicable for training the outer loop since multiple input frames are needed to construct the task-wise training samples for inner loop update.
To this end, we use Vimeo90K-Septuplet (VimeoSeptuplet) dataset~\cite{xue2018toflow}, which consists of 91,701 7-frame sequences with a fixed resolution of $448\times256$.
Though this dataset is originally designed for video super-resolution or denoising / deblocking, it is also well suited for training video frame interpolation models that require multiple frames at test time, and we train all of our models with the training split of VimeoSeptuplet dataset.
For evaluation, we use the test split of VimeoSeptuplet dataset, as well as sequences from Middlebury-\texttt{OTHERS}~\cite{baker2010flow} and HD~\cite{bao2018memc} dataset.

The \texttt{OTHERS} set from Middlebury contains 12 examples in total, with maximum resolution of $640\times480$.
We use 10 sequences with multiple input frames and remove the other two that only have two input frames and are thus not suitable for test-time adaptation.

HD dataset proposed by Bao~\etal~\cite{bao2018memc} consists of relatively high-resolution frames, from $1280\times544$ to $1920\times1080$.
Also, the length of the sequences in HD dataset is either 70 or 100, enabling test-time updates to our model.

\vspace{-3mm}
\paragraph{Implementation details}

For our experiments, we use 4 conventional video frame interpolation models as baselines: DVF~\cite{liu2017dvf}, SuperSloMo~\cite{jiang2018superslomo}, SepConv~\cite{niklaus2017sepconv}, and DAIN~\cite{bao2019DAIN}.
We first initialize each model with pre-trained parameters, provided by the authors if possible.\footnote{For SuperSloMo~\cite{jiang2018superslomo}, we use the implementations and pre-trained models from \cite{github_superslomo}.}
We denote these models as \textit{Baseline}.
Then, since we use additional training set from VimeoSeptuplet for meta-training, we also fine-tune each \textit{Baseline} models with VimeoSeptuplet training set, denoted as \textit{Re-trained} models. 
For our final \textbf{\textit{Meta-trained}} models, we start from the \textit{Baseline} model parameters and follow the iterative steps for inner and outer loop training in Algorithm~\ref{alg:meta_train}.
The reported performance for \textit{Meta-trained} models use a single inner loop update iteration at test time, and we examine the effects of increasing the number of gradient updates in the ablation study (Sec.~\ref{sec:ablation}).

We match the type of loss functions and optimization schemes for the gradient updates with the original methods used to train the \textit{Baseline} models, which differs for each method.
However, since we are fine-tuning from the pre-trained networks, we modify the inner/outer loop learning rates to be small and set $\alpha=\beta=10^{-5}$.
Throughout training, $\alpha$ is kept fixed, while $\beta$ is decayed by a factor of 5 whenever validation loss does not decrease for more than 10,000 outer loop iterations.
We do not crop patches and instead train with the full images of VimeoSeptuplet sequences with a mini-batch size of 4.
While the number of training iterations differs for each interpolation model, the full meta-training step for any model requires less than a day with a single NVIDIA GTX 1080Ti GPU since we start from the baseline pre-trained network.
The source code for our framework is made public\footnote{\url{https://github.com/myungsub/meta-interpolation}} along with the pre-trained models to facilitate reproduction.

\begin{figure*}
    \centering
	\includegraphics[width=1.00\linewidth]{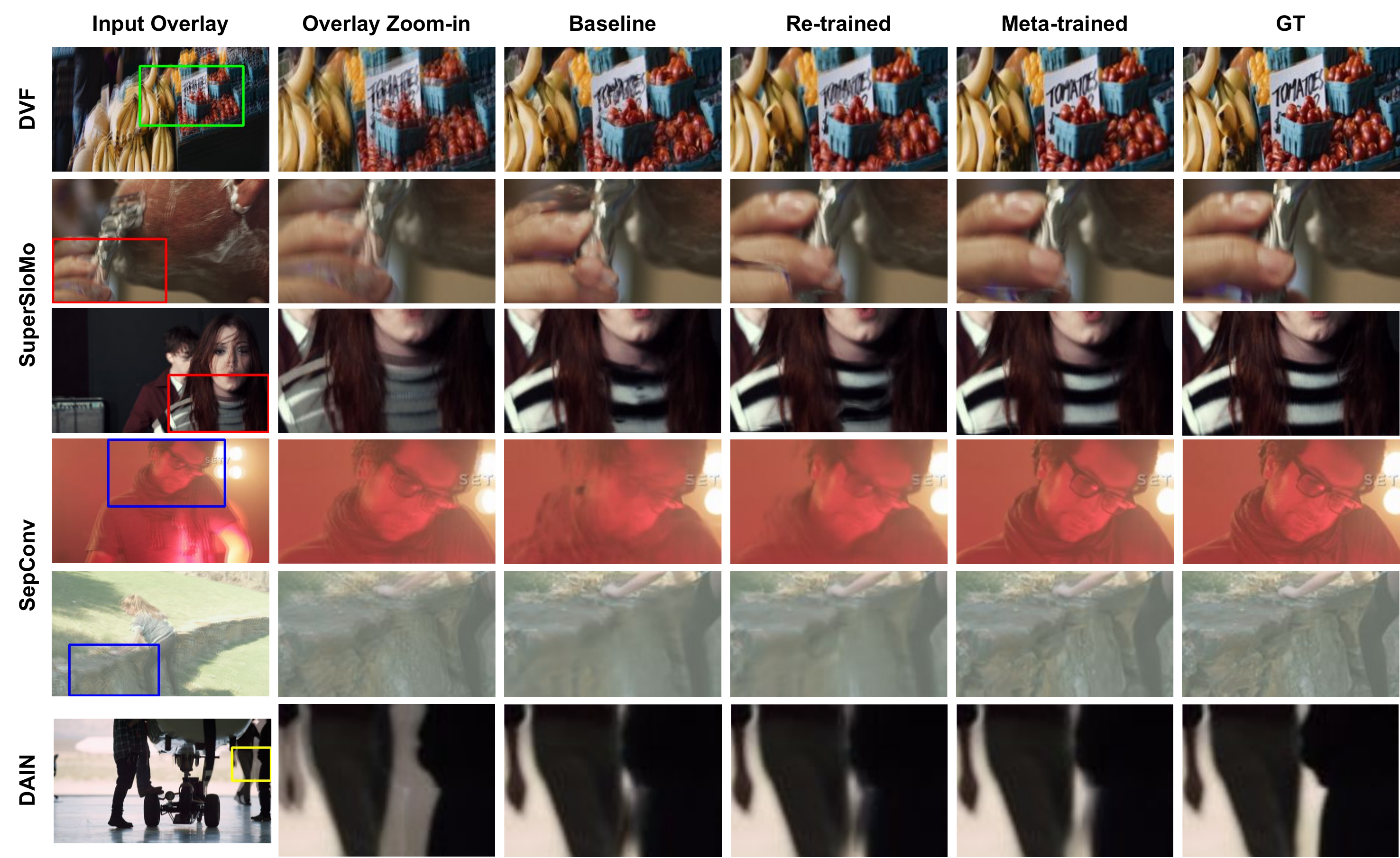}
    \caption{Qualitative results on VimeoSeptuplet~\cite{xue2018toflow} dataset for recent frame interpolation algorithms. Note how our \textit{Meta-trained} outputs infer motion substantially better than the \textit{Baseline} or \textit{Re-trained} models, as well as generate realistic textures similar to the ground truth.}
    \vspace{-2mm}
    \label{fig:qualitative}
\end{figure*}

\subsection{Video frame interpolation results}

Quantitative results for all considered baseline frame interpolation models for all evaluated datasets are summarized in Table~\ref{tb:quantitative}.
For all experiments in this section, we standardize the evaluation metric to PSNR only.
To check the results for other metrics such as interpolation error (IE) or structural similarity index (SSIM), we refer the readers to the supplementary materials.

In Table~\ref{tb:quantitative}, note the consistent performance boost achieved by the \textit{Meta-trained} model compared to both \textit{Baseline} and \textit{Re-trained} models, regardless of the method used for video frame interpolation.
Also, even though meta-training for our scene-adaptive frame interpolation algorithm is only done in VimeoSeptuplet dataset, it generalizes well to the other datasets with different characteristics, presenting the benefits of test-time adaptiveness of our approach.
Between two baselines, the \textit{Re-trained} model generally performs better than the \textit{Baseline} model.
We believe this is due to the quality (\textit{i.e.} degree of noise, artifacts, blurriness, \etc) of the training frames, since the frame sequences in VimeoSeptuplet are relatively clean.
Since DVF is trained with videos from UCF-101~\cite{soomro2012ucf} dataset that has severe artifacts, its performance increase for fine-tuning to VimeoSeptuplet was the largest.
The original training set, Adobe-240fps~\cite{su2017deep}, for SuperSloMo~\cite{jiang2018superslomo} implementation also contains some degree of noise so that re-training helps to build a stronger baseline.
An exception to this is SepConv~\cite{niklaus2017sepconv}, where re-training rather hurts the model's generalization capability to the other datasets.
Nonetheless, our \textit{Meta-trained} model considerably outperforms both baselines even for DAIN~\cite{bao2019DAIN}, the most recent state-of-the-art framework.

\begin{figure}
    \centering
    \includegraphics[width=1.0\linewidth]{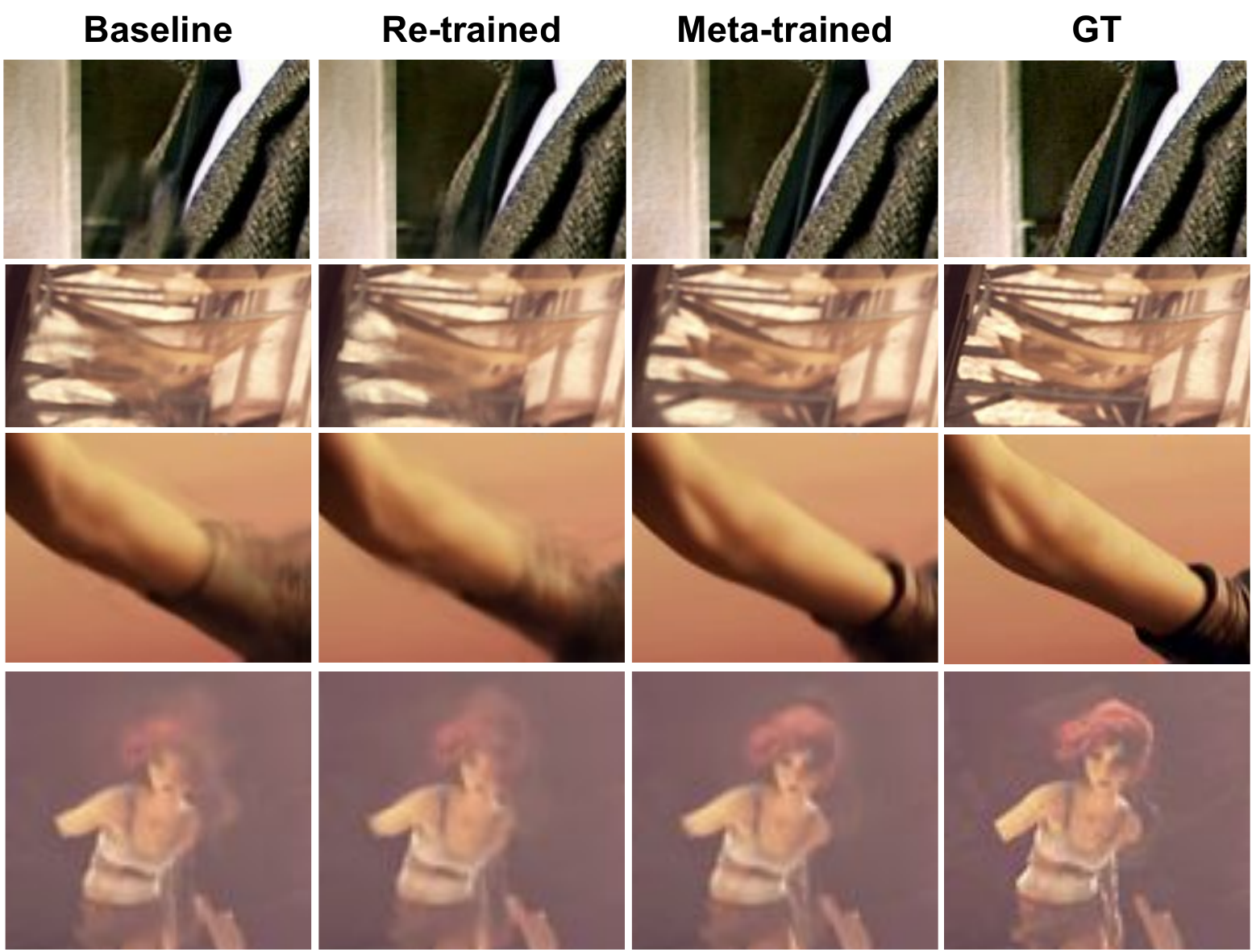}
    \caption{Qualitative results on HD~\cite{bao2018memc} dataset for SepConv~\cite{niklaus2017sepconv}. We show the cropped regions for \textit{Shields}, \textit{Alley2}, \textit{Temple2}, and \textit{Temple1} sequences.}
    \label{fig:qualitative_hd}
\end{figure}

Qualitative results for VimeoSeptuplet dataset are shown in Fig.~\ref{fig:qualitative}, where we compare the \textit{Meta-trained} model with both \textit{Baseline} and \textit{Re-trained} models for each video frame interpolation algorithm.
Note that our focus is on analyzing the benefits of \textit{Meta-trained} models with its corresponding baselines, rather than comparison between different frame interpolation algorithms.
For many cases where the baseline models fail due to large motion, our \textit{Meta-trained} model adapts to the input sequence remarkably well to synthesize better texture and more precise position of the moving regions.
In particular, the most notable improvements are shown for SepConv, which is the only model that does not utilize optical flow and the warping operation based on the predicted flow. 
Based on this evidence, we presume that explicit form of optical flow estimation constrains the possible performance gain obtainable by test-time adaptation.
Additional qualitative results for HD dataset obtained with SepConv are presented in Fig.~\ref{fig:qualitative_hd}.
Similar characteristics can be observed as in Fig.~\ref{fig:qualitative}, and our \textit{Meta-trained} model produces clearer interpolations with less artifacts.
For more qualitative comparisons and the full video demos, please see the supplementary materials.

\subsection{Ablation studies}
\label{sec:ablation}

\paragraph{Effects on the number of inner loop updates}

\begin{table}
	\centering
	\caption{Effects on varying the the number of inner loop updates. Zero updates correspond to the \textit{Re-trained} setting. PSNR (dB) for SepConv~\cite{niklaus2017sepconv} is shown for Middlebury-\texttt{OTHERS}~\cite{baker2010flow} dataset.}
	\vspace{0.2cm}
	\scalebox{0.88}{
		\begin{tabular}{l|ccccc}
		    \toprule
			\# gradient updates & 0 & 1 & 2 & 3 & 5 \\
			\midrule
			Naive Fine-tune & 34.90 & 34.90 & 34.95 & 34.99 & 35.03 \\
			Meta-trained & 34.90 & \textbf{35.81} & 35.63 & 35.58 & 35.45 \\
			\midrule
			PSNR gain & --- & \textbf{+0.91} & +0.68 & +0.59 & +0.42 \\
			\bottomrule
		\end{tabular}
	}
	\label{tb:ablation_inner_updates}
	\vspace{-3mm}
\end{table}

We vary the number of iterations for test-time adaptation and analyze the effects.
Table~\ref{tb:ablation_inner_updates} demonstrates how the final performance changes while varying the number of inner loop updates from 1, 2, 3, and 5.
We also show the results for naive test-time fine-tuning (from \textit{Re-trained} model) along with our \textit{Meta-trained} results, similar to the feasibility test in Sec.~\ref{sec:feasibility}.

In summary, meta-training for just a single inner loop update, used in most of our experiment settings, shows the most PSNR gain, while increasing the number of updates did not have any benefits on performance.
More updates even showed diminishing results, which is somewhat counter-intuitive compared to the tendency reported in MAML~\cite{finn2017model}.
We believe there are two possible reasons for this phenomenon.
First is overfitting to the data used for inner loop update ($\mathcal{D}_{{\mathcal{T}}_i}$).
In Sec~\ref{sec:feasibility}, we have shown that it is beneficial to use $\mathcal{D}_{{\mathcal{T}}_i}$ as a proxy for achieving good performance for $\mathcal{D}'_{\mathcal{T}_i}$
regardless of their distribution gap, but current ablation study suggests that \textit{over}-fitting to $\mathcal{D}_{\mathcal{T}_i}$ can have negative effects on the final performance.
This points out the need for finding the sweet spot in the trade-off between extracting from $\mathcal{D}_{\mathcal{T}_i}$ useful information that aids improving the interpolations in $\mathcal{D}'_{\mathcal{T}_i}$, and overfitting to $\mathcal{D}_{\mathcal{T}_i}$.
For video frame interpolation, an example of common useful information can be the direction of existing motion or the details on background textures.
If overfitting occurs, the inner loop may concentrate too much on handling the existing large motion and forget the generic prior knowledge learned by \textit{Baseline} pre-trained model and its \textit{Re-trained} version.
Second reason is due to growing complexity of training as the number of gradient updates increase, which makes the model susceptible to falling into local minima~\cite{finn2017model,nichol2018first}.
Presumably, incorporating recent techniques for adaptive learning rates~\cite{antoniou2018how} can help mitigate this issue, which remains as our future work.

\vspace{-3mm}
\paragraph{Effects on inner loop learning rate}
Since our algorithm starts meta-training from a pre-trained video frame interpolation model, we believe that large learning rates for the inner loop update ($\alpha$ in Algorithm~\ref{alg:meta_train}) can break the model's original performance at the early stage of training, while too small learning rates restrict the adaptive capability of the model.
To support this claim, we report the performances on setting different values of $\alpha$ in Table~\ref{tb:ablation_lr} using SepConv.
The final performance is maximized for the learning rate of $10^{-5}$, with small gaps in PSNR compared to $10^{-4}$ or $10^{-6}$.
However, regardless of the values of $\alpha$, the final performance is always better than when $\alpha = 0$, which demonstrates the effectiveness of our scene-adaptive frame interpolation algorithm via meta-learning.

\begin{table}
	\centering
	\caption{Effects on varying the learning rates for the inner loop updates. We use SepConv~\cite{niklaus2017sepconv} framework for performance comparison on VimeoSeptuplet~\cite{xue2018toflow} dataset.}
	\vspace{0.2cm}
	\scalebox{0.95}{
		\begin{tabular}{l|cccc}
		    \toprule
			Learning rate $\alpha$ & $0$ & $10^{-6}$ & $10^{-5}$ & $10^{-4}$\\
			\midrule
			PSNR (dB) & 33.72 & 34.10 & \textbf{34.17} & 34.15 \\
			\bottomrule
		\end{tabular}
	}
	\label{tb:ablation_lr}
	\vspace{-3mm}
\end{table}

\section{Conclusion}
In this paper, we introduced a novel method for video frame interpolation which aims to fully utilize the additional information available at test time.
We employ a meta-learning algorithm to train the network that can quickly adapt its parameters according to the input frames for scene-adapted inference of intermediate frames.
The proposed framework is applied to several existing frame interpolation networks and show consistently improved performance on multiple benchmark datasets, both quantitatively and qualitatively.
Our scene-adaptive frame interpolation algorithm can be easily employed to any video frame interpolation network without changing its architecture or introducing any additional parameters.

\vspace{-3mm}
\paragraph{Acknowledgements}

\noindent This work was supported by IITP grant funded by the Ministry of Science and ICT of Korea (No. 2017-0-01780), and Hyundai Motor Group through HMG-SNU AI Consortium fund (No. 5264-20190101).

{\small
\bibliographystyle{ieee_fullname}
\bibliography{egbib}

\begin{thebibliography}{10}\itemsep=-1pt

\bibitem{antoniou2018how}
Antreas Antoniou, Harrison Edwards, and Amos Storkey.
\newblock How to train your {MAML}.
\newblock In {\em ICLR}, 2019.

\bibitem{baker2010flow}
Simon Baker, Daniel Scharstein, J.~P. Lewis, Stefan Roth, Michael~J. Black, and
  Richard Szeliski.
\newblock A database and evaluation methodology for optical flow.
\newblock {\em IJCV}, 92(1):1–31, 2010.

\bibitem{bao2019DAIN}
Wenbo Bao, Wei-Sheng Lai, Chao Ma, Xiaoyun Zhang, Zhiyong Gao, and Ming-Hsuan
  Yang.
\newblock Depth-aware video frame interpolation.
\newblock In {\em CVPR}, 2019.

\bibitem{bao2018memc}
Wenbo Bao, Wei-Sheng Lai, Xiaoyun Zhang, Zhiyong Gao, and Ming-Hsuan Yang.
\newblock Memc-net: Motion estimation and motion compensation driven neural
  network for video interpolation and enhancement.
\newblock {\em arXiv preprint arXiv:1810.08768}, 2018.

\bibitem{bengio1992optimization}
Samy Bengio, Yoshua Bengio, Jocelyn Cloutier, and Jan Gecsei.
\newblock On the optimization of a synaptic learning rule.
\newblock In {\em Preprints Conf. Optimality in Artificial and Biological
  Neural Networks}, pages 6--8. Univ. of Texas, 1992.

\bibitem{carreira2017quovadis}
Joao Carreira and Andrew Zisserman.
\newblock Quo vadis, action recognition? a new model and the kinetics dataset.
\newblock In {\em CVPR}, 2017.

\bibitem{cheng2017segflow}
Jingchun Cheng, Yi-Hsuan Tsai, Shengjin Wang, and Ming-Hsuan Yang.
\newblock Segflow: Joint learning for video object segmentation and optical
  flow.
\newblock In {\em ICCV}, 2017.

\bibitem{choi2017deep}
Janghoon Choi, Junseok Kwon, and Kyoung~Mu Lee.
\newblock Deep meta learning for real-time visual tracking based on
  target-specific feature space.
\newblock {\em arXiv preprint arXiv:1712.09153}, 2017.

\bibitem{choi2020cain}
Myungsub Choi, Heewon Kim, Bohyung Han, Ning Xu, and Kyoung~Mu Lee.
\newblock Channel attention is all you need for video frame interpolation.
\newblock In {\em AAAI}, 2020.

\bibitem{danelljan2017eco}
Martin Danelljan, Goutam Bhat, Fahad Shahbaz~Khan, and Michael Felsberg.
\newblock Eco: Efficient convolution operators for tracking.
\newblock In {\em CVPR}, 2017.

\bibitem{finn2017model}
Chelsea Finn, Pieter Abbeel, and Sergey Levine.
\newblock Model-agnostic meta-learning for fast adaptation of deep networks.
\newblock In {\em ICML}, 2017.

\bibitem{glasner2009super}
Daniel Glasner, Shai Bagon, and Michal Irani.
\newblock Super-resolution from a single image.
\newblock In {\em ICCV}, 2009.

\bibitem{hochreiter2001learning}
Sepp Hochreiter, A Younger, and Peter Conwell.
\newblock Learning to learn using gradient descent.
\newblock {\em Artificial Neural Networks, ICANN 2001}, pages 87--94, 2001.

\bibitem{huang2015single}
Jia-Bin Huang, Abhishek Singh, and Narendra Ahuja.
\newblock Single image super-resolution from transformed self-exemplars.
\newblock In {\em CVPR}, 2015.

\bibitem{huang2017srhrf+}
Jun-Jie Huang, Tianrui Liu, Pier Luigi~Dragotti, and Tania Stathaki.
\newblock Srhrf+: Self-example enhanced single image super-resolution using
  hierarchical random forests.
\newblock In {\em CVPR Workshops}, 2017.

\bibitem{jiang2018superslomo}
Huaizu Jiang, Deqing Sun, Varun Jampani, Ming-Hsuan Yang, Erik Learned-Miller,
  and Jan Kautz.
\newblock Super slomo: High quality estimation of multiple intermediate frames
  for video interpolation.
\newblock In {\em CVPR}, 2018.

\bibitem{kingma2014Adam}
Diederik~P. Kingma and Jimmy Ba.
\newblock Adam: A method for stochastic optimization.
\newblock {\em CoRR}, abs/1412.6980, 2014.

\bibitem{koch2015siamese}
Gregory Koch, Richard Zemel, and Ruslan Salakhutdinov.
\newblock Siamese neural networks for one-shot image recognition.
\newblock In {\em ICML Deep Learning Workshop}, 2015.

\bibitem{liu2019cyclicgen}
Yu-Lun Liu, Yi-Tung Liao, Yen-Yu Lin, and Yung-Yu Chuang.
\newblock Deep video frame interpolation using cyclic frame generation.
\newblock In {\em AAAI}, 2019.

\bibitem{liu2017dvf}
Ziwei Liu, Raymond~A Yeh, Xiaoou Tang, Yiming Liu, and Aseem Agarwala.
\newblock Video frame synthesis using deep voxel flow.
\newblock In {\em ICCV}, 2017.

\bibitem{long2016learning}
Gucan Long, Laurent Kneip, Jose~M Alvarez, Hongdong Li, Xiaohu Zhang, and
  Qifeng Yu.
\newblock Learning image matching by simply watching video.
\newblock In {\em ECCV}, 2016.

\bibitem{meyer2018phasenet}
Simone Meyer, Abdelaziz Djelouah, Brian McWilliams, Alexander Sorkine-Hornung,
  Markus Gross, and Christopher Schroers.
\newblock Phasenet for video frame interpolation.
\newblock In {\em CVPR}, 2018.

\bibitem{meyer2015phase}
Simone Meyer, Oliver Wang, Henning Zimmer, Max Grosse, and Alexander
  Sorkine-Hornung.
\newblock Phase-based frame interpolation for video.
\newblock In {\em CVPR}, 2015.

\bibitem{michaeli2013nonparametric}
Tomer Michaeli and Michal Irani.
\newblock Nonparametric blind super-resolution.
\newblock In {\em ICCV}, 2013.

\bibitem{munkhdalai2017meta}
Tsendsuren Munkhdalai and Hong Yu.
\newblock Meta networks.
\newblock In {\em ICML}, 2017.

\bibitem{munkhdalai2018rapid}
Tsendsuren Munkhdalai, Xingdi Yuan, Soroush Mehri, and Adam Trischler.
\newblock Rapid adaptation with conditionally shifted neurons.
\newblock In {\em ICML}, 2018.

\bibitem{nam2016learning}
Hyeonseob Nam and Bohyung Han.
\newblock Learning multi-domain convolutional neural networks for visual
  tracking.
\newblock In {\em CVPR}, 2016.

\bibitem{nichol2018first}
Alex Nichol, Joshua Achiam, and John Schulman.
\newblock On first-order meta-learning algorithms.
\newblock {\em CoRR}, abs/1803.02999, 2018.

\bibitem{niklaus2018cas}
Simon Niklaus and Feng Liu.
\newblock Context-aware synthesis for video frame interpolation.
\newblock In {\em CVPR}, 2018.

\bibitem{niklaus2017adaconv}
Simon Niklaus, Long Mai, and Feng Liu.
\newblock Video frame interpolation via adaptive convolution.
\newblock In {\em CVPR}, 2017.

\bibitem{niklaus2017sepconv}
Simon Niklaus, Long Mai, and Feng Liu.
\newblock Video frame interpolation via adaptive separable convolution.
\newblock In {\em ICCV}, 2017.

\bibitem{oreshkin2018tadam}
Boris~N. Oreshkin, Pau Rodriguez, and Alexandre Lacoste.
\newblock Tadam: Task dependent adaptive metric for improved few-shot learning.
\newblock In {\em NIPS}, 2018.

\bibitem{github_superslomo}
Avinash Paliwal.
\newblock Pytorch implementation of super slomo.
\newblock \url{https://github.com/avinashpaliwal/Super-SloMo}, 2018.

\bibitem{ravi2017optimization}
Sachin Ravi and Hugo Larochelle.
\newblock Optimization as a model for few-shot learning.
\newblock In {\em ICLR}, 2017.

\bibitem{reda2019unsupervised}
Fitsum~A Reda, Deqing Sun, Aysegul Dundar, Mohammad Shoeybi, Guilin Liu,
  Kevin~J Shih, Andrew Tao, Jan Kautz, and Bryan Catanzaro.
\newblock Unsupervised video interpolation using cycle consistency.
\newblock In {\em ICCV}, 2019.

\bibitem{santoro2016meta}
Adam Santoro, Sergey Bartunov, Matthew Botvinick, Daan Wierstra, and Timothy
  Lillicrap.
\newblock Meta-learning with memory-augmented neural networks.
\newblock In {\em ICLR}, 2016.

\bibitem{schmidhuber1987evolutionary}
Jurgen Schmidhuber.
\newblock Evolutionary principles in self-referential learning.
\newblock {\em On learning how to learn: The meta-meta-... hook.) Diploma
  thesis, Institut f. Informatik, Tech. Univ. Munich}, 1987.

\bibitem{schmidhuber1992learning}
J{\"u}rgen Schmidhuber.
\newblock Learning to control fast-weight memories: An alternative to dynamic
  recurrent networks.
\newblock {\em Neural Computation}, 4(1):131--139, 1992.

\bibitem{shocher2018zero}
Assaf Shocher, Nadav Cohen, and Michal Irani.
\newblock “zero-shot” super-resolution using deep internal learning.
\newblock In {\em CVPR}, 2018.

\bibitem{simonyan2014two-stream}
Karen Simonyan and Andrew Zisserman.
\newblock Two-stream convolutional networks for action recognition in videos.
\newblock In {\em NIPS}, 2014.

\bibitem{NIPS2017_6996}
Jake Snell, Kevin Swersky, and Richard Zemel.
\newblock Prototypical networks for few-shot learning.
\newblock In {\em NIPS}, 2017.

\bibitem{soomro2012ucf}
Khurram Soomro, Amir~Roshan Zamir, and Mubarak Shah.
\newblock Ucf101: A dataset of 101 human actions classes from videos in the
  wild.
\newblock {\em CRCV-TR-12-01}, 2012.

\bibitem{su2017deep}
Shuochen Su, Mauricio Delbracio, Jue Wang, Guillermo Sapiro, Wolfgang Heidrich,
  and Oliver Wang.
\newblock Deep video deblurring for hand-held cameras.
\newblock In {\em CVPR}, 2017.

\bibitem{Sung_2018_CVPR}
Flood Sung, Yongxin Yang, Li Zhang, Tao Xiang, Philip~H.S. Torr, and Timothy~M.
  Hospedales.
\newblock Learning to compare: Relation network for few-shot learning.
\newblock In {\em CVPR}, 2018.

\bibitem{thrun2012learning}
Sebastian Thrun and Lorien Pratt.
\newblock {\em Learning to learn}.
\newblock Springer Science \& Business Media, 2012.

\bibitem{vinyals2016matching}
Oriol Vinyals, Charles Blundell, Timothy Lillicrap, koray kavukcuoglu, and Daan
  Wierstra.
\newblock Matching networks for one shot learning.
\newblock In {\em NIPS}, 2016.

\bibitem{xu2019quadratic}
Xiangyu Xu, Siyao Li, Wenxiu Sun, Qian Yin, and Ming-Hsuan Yang.
\newblock Quadratic video interpolation.
\newblock In {\em NeurIPS}, 2019.

\bibitem{xue2018toflow}
Tianfan Xue, Baian Chen, Jiajun Wu, Donglai Wei, and William~T Freeman.
\newblock Video enhancement with task-oriented flow.
\newblock In {\em CVPR}, 2018.

\bibitem{zhu2018CVPR_video_det}
Xizhou Zhu, Jifeng Dai, Lu Yuan, and Yichen Wei.
\newblock Towards high performance video object detection.
\newblock In {\em CVPR}, 2018.

\bibitem{zhu2017flow}
Xizhou Zhu, Yujie Wang, Jifeng Dai, Lu Yuan, and Yichen Wei.
\newblock Flow-guided feature aggregation for video object detection.
\newblock In {\em ICCV}, 2017.

\end{thebibliography}
}

\end{document}